\documentclass[10pt,twocolumn,letterpaper]{article}

\usepackage{wacv}
\usepackage{times}
\usepackage{epsfig}
\usepackage{graphicx}
\usepackage{amsmath}
\usepackage{amssymb}
\usepackage{color}
\usepackage{algorithm}
\usepackage{enumitem}
\usepackage[noend]{algpseudocode}
\usepackage[font=footnotesize, caption=false]{subfig}
\usepackage[percent]{overpic}
\usepackage[export]{adjustbox}
\usepackage{mathtools}
\usepackage[pagebackref=true,breaklinks=true,letterpaper=true,colorlinks,bookmarks=false]{hyperref}
\renewcommand{\etal}{ {\it et al.}}
\newcommand{\etalspace}{ {\it et al.} }

\DeclarePairedDelimiter\floor{\lfloor}{\rfloor}

\wacvfinalcopy 

\def\wacvPaperID{330} 

\ifwacvfinal\fi
\begin{document}

\title{Segmenting root systems in X-ray computed tomography images using level sets}

\author{Amy Tabb\\
USDA-ARS-AFRS\\
Kearneysville, West Virginia, USA\\
{\tt\small amy.tabb@ars.usda.gov}
\and
Keith E. Duncan, Christopher N. Topp \\
Donald Danforth Plant Science Center\\
St. Louis, Missouri, USA \\
{\tt\small kduncan, ctopp@danforthcenter.org}
}

\maketitle
\ifwacvfinal\thispagestyle{empty}\fi

\newcommand{\diver}{\mathop{\mathrm{div}}\nolimits}

\begin{abstract}

The segmentation of plant roots from soil and other growing media in X-ray computed tomography images is needed to effectively study the root system architecture without excavation.  However, segmentation is a challenging problem in this context because the root and non-root regions share similar features.  In this paper, we describe a method based on level sets and specifically adapted for this segmentation problem.  In particular, we deal with the issues of using a level sets approach on large image volumes for root segmentation, and track active regions of the front using an occupancy grid.  This method allows for straightforward modifications to a narrow-band algorithm such that excessive forward and backward movements of the front can be avoided, distance map computations in a narrow band context can be done in linear time through modification of Meijster et al.'s distance transform algorithm, and regions of the image volume are iteratively used to estimate distributions for root versus non-root classes.  Results are shown of three plant species of different maturity levels, grown in three different media.  Our method compares favorably to a state-of-the-art method for root segmentation in X-ray CT image volumes. \footnote{The citation information for this paper is: A. Tabb, K.E. Duncan, and C.N. Topp, “Segmenting root systems in X-ray computed tomography images using level sets”, 2018 IEEE Winter Conference on Applications
of Computer Vision (WACV), Lake Tahoe, NV/CA. pp. 586-595. DOI 10.1109/WACV.2018.00070} \footnote{Mention of trade names or commercial products in this publication is solely for the purpose of providing specific information and does not imply recommendation or endorsement by the U.S. Department of Agriculture.  USDA is an equal opportunity provider and employer.}\footnote{This material is based upon work supported in part by the U.S. National Science Foundation under Award number IOS-1638507 to C.N. Topp, and by a Cooperative Agreement between Valent BioSciences LLC and C.N. Topp.} 
\end{abstract}

\section{Introduction}
\label{sec:init}

The phenotyping of three-dimensional root system architecture traits is necessary for effective genetic studies that will identify connections between root traits and their genotypic control, which in turn may unlock the potential of roots for crop improvement.  However, the measurement of these traits proves difficult; unlike in above-ground work on shoots where manual measurements can be made, any manual measurement of root system architecture (RSA) traits results in destruction of the plant because it is displaced from the soil. That being said, it remains that for many contexts excavation for manual measurement, the `shovelomics' method, still remains the most reliable method for RSA \cite{Abiven2015, Colombi2015, Trachsel2011}.

Other approaches include growing plants in gels \cite{clark2011three, Zheng2011Detailed} and hydroponic systems \cite{pineros2016evolving}; the root shape is extracted using a turntable system with a shape from silhouette approach.  These systems allow the plant to continue growing, and its root system to be examined over time \cite{Symonova2015dynamic, Topp2013-3D}.  One concern with acquiring RSA traits from plants grown in gels or liquids in that these materials influence the RSA in ways that soil does not  \cite{pineros2016evolving}, and there are constraints to using gel-based systems, such as the size and maturity of the plant.  In addition, the frictional and nutritional properties of the gel growing medium does not exactly simulate the environment plant roots experience in soil-based media.

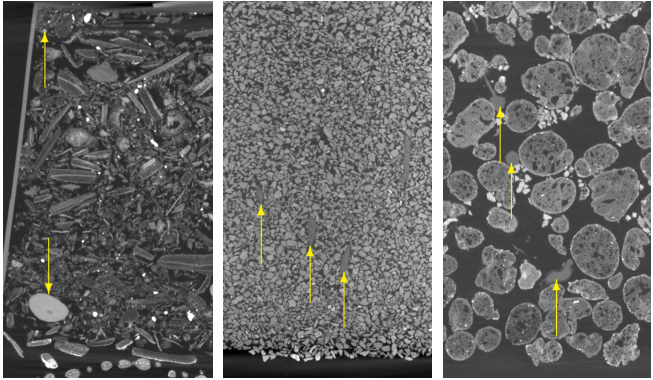
\begin{figure}
\subfloat[Cassava in a potting mix] 
{
	\begin{overpic}[ height ={5cm}, trim={0 0 580 0}, clip]{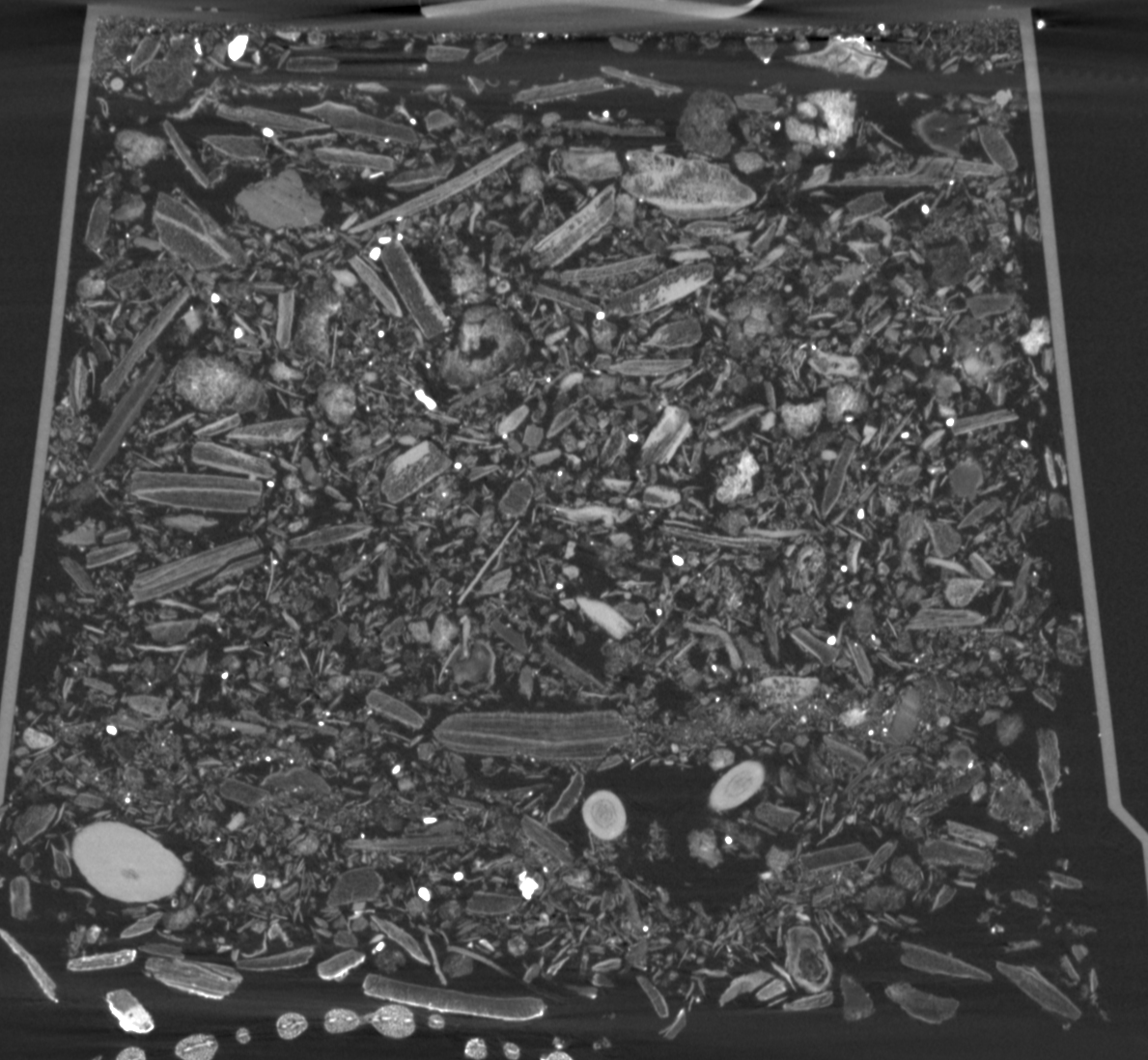}
	\put(12,37){\color{yellow}\vector(0,-1){15}}
	\put(11,77){\color{yellow}\vector(0,1){15}}
	\end{overpic}
}
\subfloat[Maize in turface] 
{
	\begin{overpic}[height ={5cm}, trim={320 0 200 0}, clip]{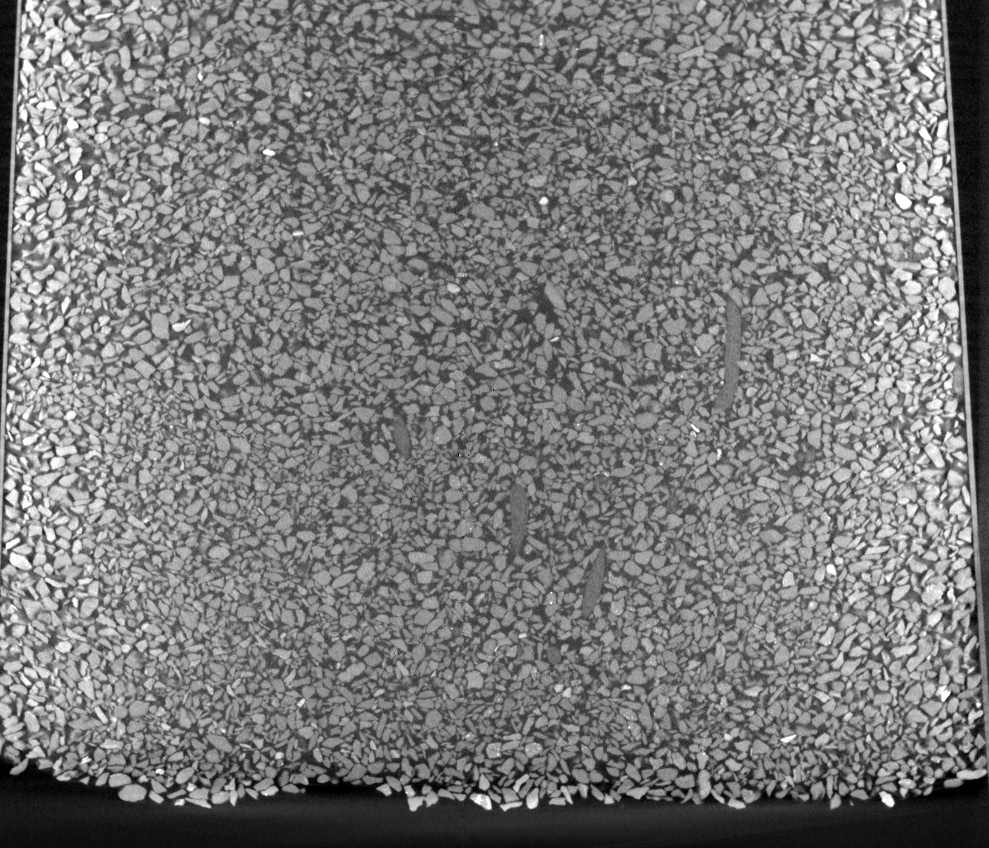}
	\put(10,30){\color{yellow}\vector(0,1){16}}
	\put(23,20){\color{yellow}\vector(0,1){15}}
	\put(32,13){\color{yellow}\vector(0,1){15}}
	\end{overpic}
}
\subfloat[Soybean in expanded clay] 
{
	\begin{overpic}[height ={5cm}, trim={400 0 450 0}, clip]{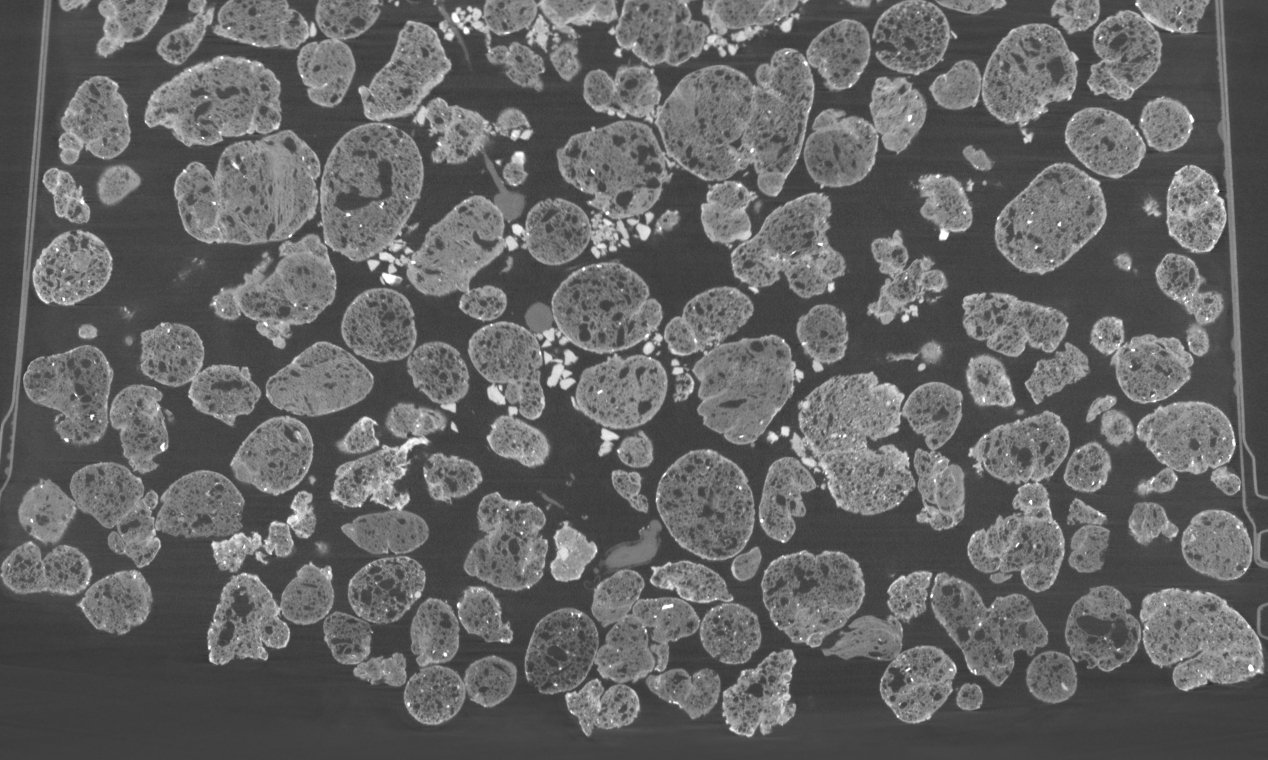}
	\put(30,11){\color{yellow}\vector(0,1){15}}
	\put(18,42){\color{yellow}\vector(0,1){15}}
	\put(15,57){\color{yellow}\vector(0,1){15}}
	\end{overpic}
}
\caption{{\bf Best viewed in color.} Example of X-ray CT images from datasets 3, 5, and 6 from this paper, with details given in Table \ref{tab:DatasetInfo}, to show the variety of medium and root system shapes.  Some root regions are indicated with a yellow arrow; not all root regions in each image are indicated. }
\end{figure}

For these reasons, the use of X-ray computed tomography (CT) has emerged as a way to visualize the structure of actively-growing plant root systems in soil or soil-like media \cite{Flavel2017image, Mairhofer2013Recovering, Mairhofer2012Rootrak, Metzner2015direct}, with the possibility of capturing root growth over time because the technique is nondestructive.  In order to quantify the root's shape, the root regions of the X-ray CT regions must be segmented from the non-root regions, which is extremely arduous to do manually for complex root systems.  Once the segmentation is complete, features of the root systems can be measured using metrics within the root community and this data is then used for various physiological and genomic studies. 

In this paper, we provide a three-dimensional level set method adapted to the root segmentation problem in X-ray CT, specifically when the volumes are large ($1803 \times 2238 \times 1805$, or $7.2 \times 10^9$ voxels).  Our method requires minimal, non-exact initialization of a selection of image slices, of various orientations, from the X-ray CT volume.  The level set formulation is derived from Rousson and Deriche \cite{Rousson2002Variational_TR, Rousson2002Variational_WS}, but instead of modeling the whole CT volume, only regions close to the root class voxels are represented by the root and non-root distributions.  We introduce some modifications specific for dealing with root systems as well as the size of the datasets we encounter.  Using a narrow-band approach to level sets, we denote active regions of the contour using an occupancy grid, where the dimension of each grid cube edge is at least the size of the band.  This grid is used for the updating of the distributions as well as the distance map computations.  Each voxel is permitted to be a member of the active region of the contour until a threshold is exceeded, at which point the voxel becomes a permanent member of the static contour.  This process restricts computational resources to the active region, and since the roots are thin, prevents previously discovered root regions from being smoothed away. The method is demonstrated on five datasets, with three plant species: soybean ({\it Glycine max}), maize ({\it Zea mays}), and cassava ({\it Manihot esculenta}) of various different maturity levels and root thicknesses, and grown in three different media: expanded clay, turface, and potting mix.  Specifically, our contributions are:

\begin{itemize}[noitemsep]
\item{General algorithm to segment root systems from non-root regions from a range of media with a varied texture and size characteristics in X-ray CT volumes.}
\item{A level set method that focuses computational resources on the active region of the contour through use of an occupancy grid and contour age parameter.}
\item{Fast exact narrow band distance map computations within the positive occupancy grid regions by an alteration of Meijster \etalspace \cite{meijster2002general} distance transform algorithm.}
\end{itemize}

Level sets have been in use for some time in the medical community to segment items of interest in the human body for quantification or further review from CT image volumes, such as skulls, lung nodules, and portions of the heart (\cite{Ghadimi2016skull, Farag2013novel, Shahzad2017automated, Yang2017left} respectively).  In the medical community, segmentations are often desired in a clinical setting, so one concern is the computational time involved with level set implementations. Lefohn \etalspace \cite{lefohn2004streaming} implemented level sets on the GPU to mitigate this concern.  Wang \etalspace \cite{Wang2011level, Wang2014fast} proposed a front evolution strategy called coherent propagation, in which the oscillation of the front is reduced resulting in reduced computational time. The coherent propagation strategy was implemented in a bone segmentation setting in \cite{Chowdhury2017cortical}.

The prior work on segmenting X-ray CT images of roots fall into two groups: those based on threshold and/or morphological techniques within established software packages  \cite{Flavel2017image, Metzner2015direct} and those that use a two-dimensional level set formulation to track root regions throughout the scan volume  \cite{Mairhofer2013Recovering, Mairhofer2012Rootrak}.  All of the prior work operations on X-ray micro-computed tomography.  The method described in Metzner \etalspace 2015 \cite{Metzner2015direct} specifies thresholds for small, medium, and large roots, and then uses connected components and morphological operations to locate root regions and a filter to remove air or non-root regions.  Flavel \etalspace 2017 \cite{Flavel2017image} created a plugin for ImageJ, Root1, that normalizes the greylevels between two scans, and then performs adaptive thresholding, filtering and morphological operates to locate the root regions. Mairhofer \etalspace 2012 with RooTrak \cite{Mairhofer2012Rootrak} use a level-set variation to track roots from the top of the pot to the bottom.  This is done by the user initializing the root regions on the top layer of the image volume.  Then, the segmentation at the current layer is determined using a level set method.  Then, the subsequent layer down is initialized with the segmentation from the current layer, and the segmentation into root and non-root regions is again found, and so on until the bottom layer.  In \cite{Mairhofer2013Recovering}, the authors supplemented the system with the ability to recover upward-growing roots as well.  Finally, the key ideas of RooTrak were extended in  \cite{Mairhofer2016Visual} to track, and differentiate, two or more plants whose roots are growing in the same pot.  

This work is most similar to Mairhofer \etal's RooTrak \cite{Mairhofer2013Recovering, Mairhofer2012Rootrak} in that both employ level set formulations.  However, our method incorporates the three-dimensional growth of root systems in a different way by directly implementing a three-dimensional level set segmentation algorithm.  In addition, we allow for different kinds of orientations and initializations.  

\section{Background and related work on level sets}
\label{sec:background}

We formulate the segmentation problem using a level set approach, following the general form of Rousson and Deriche 2002 \cite{Rousson2002Variational_TR, Rousson2002Variational_WS} in their formulation for scalar images (a review of level sets for segmentation  is \cite{Cremers2007}, and much of the basic notation here is consistent with that work).  Let $\Omega$ be the image domain, and in this paper $\Omega$ consists of all the pixels forming the images of the X-ray CT volume.  The segmentation problem is to find the division of $\Omega$ into two sets $\Omega_1$ and $\Omega_2$.  The boundary dividing the two sets of pixels is the contour $\mathcal{C}$.  Using the implicit representation of a contour, the set of pixels $\mathcal{C}$ is the zero level set of the embedding function $\phi: \Omega \mapsto \mathbb{R}$, so that $\mathcal{C} = \lbrace x \in \Omega | \phi(x) = 0 \rbrace$.  $\phi$ then is the signed distance function from the contour: $0$ at the contour, positive for pixels in $\Omega_1$ and negative for pixels in $\Omega_2$.  

In the Rousson and Deriche formulation, it is assumed that the two classes of pixels are drawn from Gaussian distributions, $\mathcal{N}(\mu_1, \sigma_1^2)$, $\mathcal{N}(\mu_2, \sigma_2^2)$. Given an initial division of pixels into $\Omega_1$ and $\Omega_2$, with Gaussian parameters $\theta_1$ and $\theta_2$, the energy function to minimize is
\begin{equation}
E(\Omega_1, \Omega_2) = -\sum \int_{\Omega_i} \log p_i \left(x | \theta_i \right) \mathop{dx} + \nu |\mathcal{C}|
\label{eq:energy}
\end{equation}
where $x$ is the greyvalue of a pixel and $\nu$ is a regularization parameter that favors smaller contour surfaces $|\mathcal{C}|$.

The gradient descent equation to evolve $\phi$ is
\begin{multline}
\frac{\mathop{d\phi}}{\mathop{dt}} = \delta \left( \phi \right) \Bigl( \nu \diver  \Bigl( \frac{\mathop{\nabla \phi}}{|\mathop{\nabla \phi|}}\Bigr)\\
 + \frac{\left(I-\mu_2  \right)^2}{2\sigma_2^2} - \frac{\left(I-\mu_1  \right)^2}{2\sigma_1^2} + \log\frac{\sigma_2}{\sigma_1}\Bigr)
\label{eq:gradient}
\end{multline}

The process in many level-set methods, per iteration, given these preliminaries can be represented roughly in four steps as shown in Algorithm \ref{alg:overall}. Because of the presence of $\delta \left( \phi \right)$ in Equation \ref{eq:gradient}, we use the narrow band hypothesis, and the band distance is a parameter in our method.  To compute the curvature $\diver  \Bigl( \frac{\mathop{\nabla \phi}}{|\mathop{\nabla \phi|}}\Bigr)$, we use the difference of normals method \cite{Whitaker2001variable, lefohn2004streaming}, and as in \cite{Rousson2002Variational_TR} approximate $|\phi| = 1$ because $\phi$ is the signed distance function.

\begin{algorithm}
\caption{General algorithm for front evolution for one iteration of level set segmentation} \label{alg:overall}
\begin{algorithmic}[1]
\State {Locate $\mathcal{C}$ and update signed distance map $\phi$.}\label{step1}
\State {Estimate current parameter states $\theta_1 = \mathcal{N}(\mu_1, \sigma_1^2)$, $\theta_2 = \mathcal{N}(\mu_2, \sigma_2^2)$ from current division of $\Omega_1$, $\Omega_2$.} \label{step2}
\State {Evolve the level set by computing $\frac{\mathop{d\phi}}{\mathop{dt}}$, Eq. \ref{eq:gradient}.} \label{step3}
\State {$\phi^{\left( i+1\right)} = \phi^{\left( i\right)} +  \frac{\mathop{d\phi}}{\mathop{dt}}$}\label{step4}
\end{algorithmic}
\end{algorithm}

\section{Level sets for segmenting root systems}

We modified many aspects of the approach presented in \cite{Rousson2002Variational_TR, Rousson2002Variational_WS} such that segmentations of large volumes with small, thin roots would be successful.  There are two major ideas, and both are implemented through the use of occupancy grids.  The first idea is that the computational resources should be focused on the active portions of the contour (Section \ref{ss:mod}).  The second idea is that the image domain $\Omega$ is not divided into two sets $\Omega_1$ and $\Omega_2$, but rather three: $\Omega_1$,  $\Omega_2$, and $\Omega_U$, and by $\Omega_U$ we indicate those voxels that have not been explored yet and are unlabeled.  By doing so, we are able to take into account some of the physical characteristics of the root CT volumes (Section \ref{ss:g-h}).  Before we get into the details of these ideas and how they affect the steps presented in Section \ref{sec:background} of the Rousson and Deriche level set implementation, we describe the initialization procedure in the context of root segmentation in Section \ref{ss:init}.    

\subsection{Initialization}
\label{ss:init}

\begin{figure*}[]
\centering
\subfloat[Cassava in potting medium] 
{
\label{sf:cassava_init}
	\includegraphics[width=0.32\linewidth]{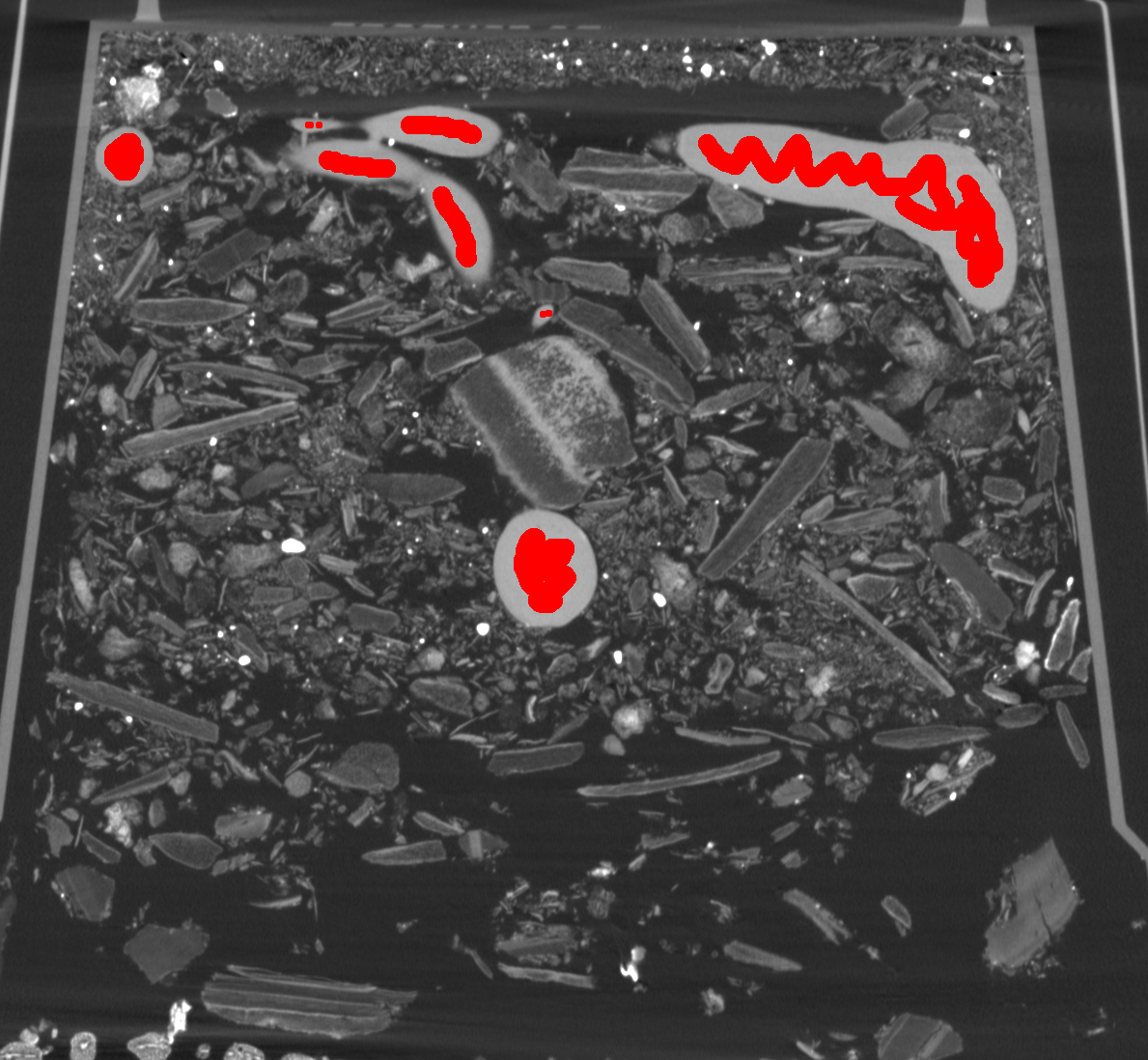}
}
\subfloat[Maize in turface] 
{
\label{sf:maize_init}
	\includegraphics[width=0.32\linewidth]{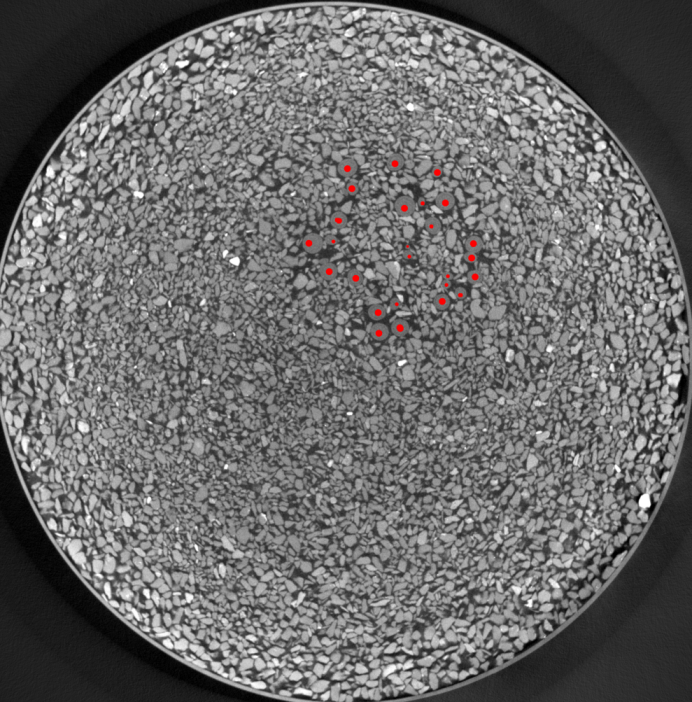}
}
\subfloat[Maize in expanded clay] 
{
\label{sf:maize_ec_init}
	\includegraphics[width=0.32\linewidth]{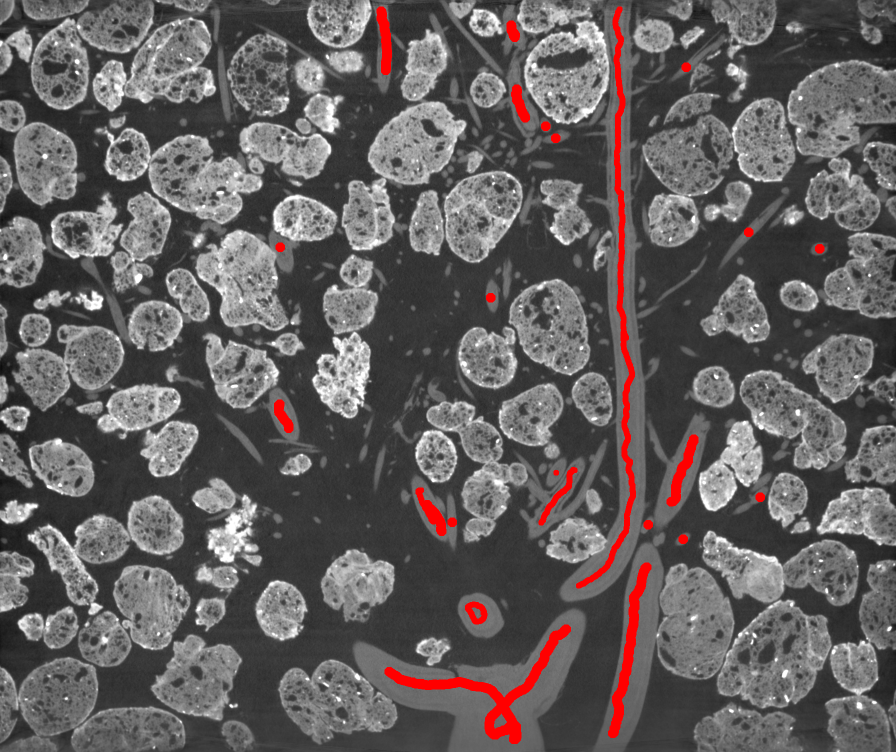}
}
\caption{\textbf{Best viewed in color.} Example of X-ray CT initialization images.  Images are marked by the user indicating root regions.  We used the color red.  \ref{sf:cassava_init} and \ref{sf:maize_ec_init} show image slices whose orientation are perpendicular to the ground plane, while \ref{sf:maize_init} shows an image slice whose orientation is parallel to the ground plane.}
\label{fig:init}
\end{figure*}

As mentioned in Section \ref{sec:background}, $\Omega_1$ and $\Omega_2$ are initialized. In the context of root segmentation, this must be done manually.  In \cite{Mairhofer2013Recovering, Mairhofer2012Rootrak}, the user initializes the root regions from the top layer of the image volume, corresponding to the top layer of the material in the pot.  In our system, the user initializes a selection of images by marking the root regions, which will make up $\Omega^0_2$, with red color. Examples are shown in Figure \ref{fig:init}.  The choice of red is arbitrary, and can be changed to better match user preferences.  Since these regions will serve as the initial values of the root class segmentation, it is important that only roots are marked, but it is not necessary that all roots are marked in an image.  Our practice is to indicate root regions with squiggles or dots on the larger root regions, and leave regions where we are unsure about the presence or absence of a root unmarked.  In particular, in the turface  medium, it can be difficult for the user to determine root versus non-root regions (Figure \ref{sf:maize_init}).

The orientation of initialization images is chosen according to the ease of locating the root region for that particular plant and medium.  For example, in maize the plant has a great deal of downward growth, so it may be easier to localize roots in image slices from the CT volume that are parallel to the ground plane than in other orientations, particularly for younger plants, as in Figure \ref{sf:maize_init}.  On the other hand, more mature plants, such as that shown in Figure \ref{sf:maize_ec_init} have large root regions that may be initialized from slices perpendicular to the ground plane, as with plants that grow in a variety of directions, such as cassava (Figure \ref{sf:cassava_init}).  Multiple images are marked per dataset, possibly from multiple orientations as appropriate, with the number of images being roughly proportional to the complexity of the root system or dataset size.  Since we use the narrow band assumption, the distance from the initialization regions and the root regions to be discovered will influence the number of iterations of steps 1-4 in Section \ref{sec:background}.  For this reason, we typically space the initialization images evenly throughout the volume.  

\subsection{Active regions of the contour}
\label{ss:mod}

\begin{figure}[]
\centering
\includegraphics[width=0.65\linewidth]{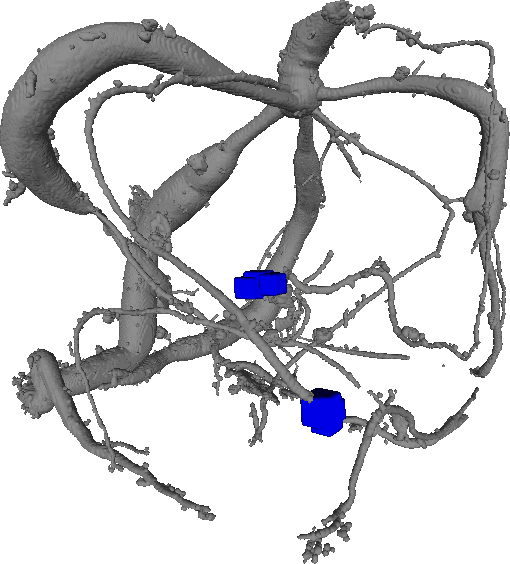}
\caption{An illustration of active grid regions relative to the static portions of the contour, Dataset 2.   The occupancy grid $G_a(y) = 1$ regions are represented by the blue squares, at iteration $200$.  The portions of the root structure shown, in gray, represent $\mathcal{C}_s$.  The number of occupancy grid cubes where $G_h(y) = 1$ is $1310904$, whereas the number of occupancy grid cubes where $G_a(y) = 1$ is 147.  More details about the results of this experiment are shown in Table \ref{tab:DatasetInfo}.}
\label{fig:dataset2composite}
\end{figure}
The modifications we make to the general level set algorithm in Algorithm \ref{alg:overall} has at its base the splitting of the contour $\mathcal{C}$ into two parts: those regions representing the active region of the contour $\mathcal{C}_a$ and those representing the static regions of the contour $\mathcal{C}_s$. This splitting operation is done for two reasons: 1) to reduce runtime, 2) prevent root regions discovered in early iterations from being smoothed away in later iterations.  Others have pursued strategies that aim to set portions of the contour with the aim of reducing runtime of level set methods, such as the coherent propagation strategy \cite{Wang2011level, Wang2014fast} mentioned in Section \ref{sec:init}.  In this section, we will explain our methodology.
 
A voxel is classified as part of one or the other contour sets in the following way.  Given that the algorithm is at iteration $i$, a function $count\left(x\right)$ records the number of times a voxel has been a member of $\Omega_2$ (the root class) from iteration $0$ to $i$.  A parameter $t$ represents the maximum value of $count\left(x\right)$ such that voxel $x$ is part of $\mathcal{C}_a$.  Consequently, we define the two sets according to this function, $t$, and $\mathcal{C}$ for any voxel $x$: $\mathcal{C}_a = \{x | count\left(x\right) \leq t \cap x \in \mathcal{C}\}$ and $\mathcal{C}_s = \mathcal{C} \setminus \mathcal{C}_a$.  Figure \ref{fig:dataset2composite} shows an example of the $G_a(y) = 1$ regions versus the $\mathcal{C}_s$ regions.

With this definition of the active region of the contour, we will now the introduce our use of occupancy grids into the level set method.  Let $G_a$ be the occupancy grid indicating the presence or absence of active regions of the contour $\mathcal{C}_a$.  Each cube of $G_a$ covers $s \times s \times s$ voxels, where $s \geq b$ and $b$ is the band size for the narrow-band distance computation of $\phi$ (Line \ref{step1} from Algorithm \ref{alg:overall}).  To mark $G_a$ per iteration, first all entries are $0$. Given a voxel in $\mathcal{C}_a$, the grid cube location is found and marked $1$, and its 26-connected neighbors within the grid coordinate system are found, and also marked $1$.  This is repeated for every voxel in $\mathcal{C}_a$. The result is that the grid cubes marked $1$ are guaranteed to contain the tube of voxels within $b$ voxels of $\mathcal{C}_a$, provided $s \geq b$. 

$G_a$'s guarantee of containment of the $b$-voxel wide tube of voxels around $\mathcal{C}_a$ is important, because we use an exact linear time algorithm for computing the distance map $\phi$ only within this region. Use of such an algorithm provides an efficient narrow-band distance transform implementation. The algorithm we use for distance transform (DT) computation is that of Meijster \etalspace \cite{meijster2002general}, a grid sweep algorithm whose complexity is linear in the number of elements in the grid.  Another advantage of the algorithm in \cite{meijster2002general} is that it is easily parallelizable. Suppose the smallest Euclidean distance from voxel $v$ to the contour is $EDT_v$.  Then, the truncated Euclidean distance is $TEDT_v = \min(b+1, EDT_v)$, so that all voxels within the narrow band will have $TEDT_v \leq b$.  We alter \cite{meijster2002general} to calculate the truncated Euclidean distance, and this altered version is applied within the regions of the image volume where $G_a(y) = 1$.  We notate this set of voxels as $V$, or $V = \lbrace x | x \in G_a(y) \wedge G_a(y) = 1\rbrace$. As a result, the complexity of our distance transform step is linear in $|V|$.  

The modification of the process is summarized in Algorithm \ref{alg:mod_step1} for one iteration.  First, all entries of $G_a$ are set to zero.  Line \ref{mod:line1} is as before from Algorithm \ref{alg:overall}.  The active and static portions of the contour are determined in lines \ref{mod:line2}-\ref{mod:line3}.  Then, grid cubes of $G_a$ that include active regions of the contour are updated in lines \ref{mod:line4}-\ref{mod:line7}.  The modified distance transform of Meijster \etalspace is applied to those voxels in $V$ in \ref{mod:line9}.  We note that $\phi$ is a signed distance transform; so voxels $x \in \Omega_2$ have $\phi_x = -TEDT_x$. For those voxels within the narrow band, the gradient and new value of $\phi_x$ is computed (lines \ref{mod:line11}-\ref{mod:line13}).  We compute distributions by use of histograms of the image values; if the class labels change, we need to update the histogram with the change (line \ref{mod:line15}).  Finally, updated distributions can be quickly computed by way of examining the histograms for each class (line \ref{mod:line16}).

\begin{algorithm}
\caption{Modification of general algorithm for root segmentation} \label{alg:mod_step1}
\begin{algorithmic}[1]
\Require{$\forall y \quad  G_a\left(y\right) = 0$}
\State {Locate $\mathcal{C}$ from $\phi$} \label{mod:line1}
\State {$\mathcal{C}_a = \{x | count\left(x\right) \leq t \cap x \in \mathcal{C}\}$}\label{mod:line2}
\State {$\mathcal{C}_s = \mathcal{C} \setminus \mathcal{C}_a$}\label{mod:line3}
\For   {$x \in \mathcal{C}_a$}\label{mod:line4}
\State { $[xc, yc, zc]$= coordinates of $x$ in image volume }\label{mod:line5}
\State { $[\floor{xc/s}, \floor{yc/s}, \floor{zc/s}]$= $gc$}\label{mod:line6}
\State {Mark $G_a\left(gc\right) = 1$ and all 26-connected neighbors of $gc$ in $G_a$ to $1$.}\label{mod:line7}
\EndFor
\State {$V = \lbrace x | x \in G_a(y) \wedge G_a(y) = 1\rbrace$}\label{mod:line8}
\State {$\forall x \in V$, update $\phi$ using modification of DT in \cite{meijster2002general}.}\label{mod:line9}
\For {$x \in V$}\label{mod:line10}
\If {$|\phi_x| \leq b$}\label{mod:line11}
\State {Compute $\frac{\mathop{d\phi_x}}{\mathop{dt}}$, Eq. \ref{eq:gradient}.}\label{mod:line12}
\State {$\phi_x^{\left( i+1\right)} = \phi_x^{\left( i\right)} +  \frac{\mathop{d\phi_x}}{\mathop{dt}}$} \label{mod:line13}
\If {$sign(\phi_x^{\left( i+1\right)}) \neq sign(\phi_x^{\left( i\right)})$}\label{mod:line14}
\State {Update class histograms with class changes.} \label{mod:line15}
\EndIf
\EndIf
\EndFor
\State {Estimate $\theta_1 = \mathcal{N}(\mu_1, \sigma_1^2)$, $\theta_2 = \mathcal{N}(\mu_2, \sigma_2^2)$ from histograms.} \label{mod:line16}
\end{algorithmic}
\end{algorithm} 

\subsection{Occupancy grid for gradual exploration of image volumes}
\label{ss:g-h}

Large pots are characterized not only by correspondingly large image volumes, but also variations within the pot as a result of compaction, water density variation, and root density variation.  The initialization procedure outlined in Section \ref{sec:init}, which produces a set $\Omega^0_2$, and following the classical procedure of dividing $\Omega$ into two classes, may not produce a set $\Omega^0_1$ with distribution parameters $\Theta^0_1$ that are representative of the regions surrounding $\Omega^0_2$. To deal with this issue, we also use the occupancy grid $G_a$ introduced in Section \ref{ss:mod} to incrementally add new regions to $\Omega_1$.  To do so, we require a second occupancy grid $G_h$, where the grid cubes where $G_h(y) = 1$ represents regions that have been explored, or the history of the regions that the method has visited in the evolution of the front.

The algorithm for this procedure is shown in Algorithm \ref{alg:G-h}.  Any grid entries where $G_a\left(y\right) = 1$ are checked against $G_h(y)$; if $G_h(y)$ has already been explored, or $G_h(y) = 1$, then nothing is updated.  Otherwise, $G_h(y)$ is set to explored, and its contents are added to the class histogram for $\Omega_1$, non-root regions.  Grid cubes where $G_h(y) = 0$ represent $\Omega_U$, or the unlabeled region.

Note that we set the same grid edge length $s$ for $G_a$ and $G_b$, but that it is possible to alter this scheme by making one grid cube edge length a multiple of the other, for instance, and the implementation is still straightforward.  One may want to pursue such a strategy to reflect some differences between the band size for the narrow band implementation of level sets and the assumptions about the distance around root regions ($\Omega_2$) needed to properly model it.

\begin{algorithm}
\caption{Updating of $G_h$ and class histogram of $\Omega_1$} \label{alg:G-h}
\begin{algorithmic}[1]
\For{$\forall y $ where $ G_a\left(y\right) = 1$}
\If {$G_h(y) = 0$}
\State {Add voxels in $G_h(y)$ to class histogram for $\Omega_1$.}
\State {$G_h(y) = 1$}
\EndIf
\EndFor
\end{algorithmic}
\end{algorithm} 

\subsection{Termination criterion and extraneous region removal}

The termination of the method occurs when $\mathcal{C}_a < k$, where $k$ is a constant.  In other words, the number of voxels on the active portion of the contour has become small enough that we terminate.

Following termination of the iteration level set procedure, the connected components are determined from the set of the voxels that are in $\Omega_2$.  Those components that are connected to the initialization set $\Omega^0_2$ are retained, the voxels in the remaining components are marked as $\Omega_1$, the non-root class.

\section{Experiments}

X-ray CT images were acquired by a North Star Imaging (NSI) X5000 X-ray CT instrument.  Five datasets were acquired with a range of characteristics, and they are listed in Table \ref{tab:DatasetInfo}.  In some cases, only a selection of the image slices were selected from the entire volume because the root system had not yet grown through the entire pot.  In this case, the truncated volume dimensions are listed in the table.

\begin{table*}
\centering{}\caption{\label{tab:DatasetInfo}Information about the X-ray CT datasets used in the experiments, as well as parameters used to run the method described in this paper.  The eighth column designates the number of initialization images used, while the ninth shows the number of iterations of the algorithm until convergence.  The final column gives the method's total runtime, excluding the image load time.}
\resizebox{\linewidth}{!}
{
\begin{tabular}{|l|ccccccccc|}
\hline
Dataset & Plant species & Medium & Voxel size & Dimensions & $b$ & $\nu$ & No. init. images & No. iter.s & Time\\ \hline
1 & Maize & expanded clay & 66 $\mu$m & $1792 \times 1504 \times 1792$ & 10 & 1 &  6  & 883 & 4.79 hr.s \\ \hline
2 & Cassava & Berger 45 potting medium & 55 $\mu$m  & $1189 \times 1019 \times 1079$ & 10 & 1 & 11 & 302 & 25.26 min.s \\ \hline
3 & Soybean  & expanded clay &  62 $\mu$m  & $2536 \times 1399 \times 251$  & 10 & 1.2 & 1 & 171 & 8.03 min.s \\ \hline
4 & Maize & expanded clay & 100  $\mu$m & $1803 \times 2238 \times 1805 $ & 20 & 1.5 & 17 & 514 & 4.69 hr.s \\ \hline
5 & Maize & turface & 1 mm &  $1977 \times 1626 \times 1801$ & 10 & 1.05 & 19 & 598 & 3.02 hr.s \\ \hline
\end{tabular}
}
\end{table*}

\subsection{Implementation details}

The parameters described in the method are also shown in Table \ref{tab:DatasetInfo}. The maximum contour membership threshold $t$ was set to $1$ for all datasets, and the termination constant $k=100$ for all datasets as well.  The absence of material, such as outside of a pot, manifests in the images as a dark color.  We specify a minimum greylevel value for members of $\Omega_1$ and $\Omega_2$ to prevent voxels from being added to either set if their greylevel is below the minimum.  In addition, we specify a wide band of values that $\Omega_2$, the root class, may take.  A voxel whose greylevel is outside of this band is not permitted to change its class label from $\Omega_1$ to $\Omega_2$.  Since the root and non-root colors have such a high degree of overlap, this simple implementation detail prevents erroneous front evolutions which can be easily predicted {\it a priori}. We implemented our method in C/C++, and all results in this paper were produced on a machine with a 12 core Intel Xeon(R) 2.7 GHz processor and 256 GB RAM
 
\subsection{Results and discussion}

\begin{figure}[]\centering
\subfloat[Side view] 
{
\label{sf:maize0}
\includegraphics[width=0.5\linewidth]{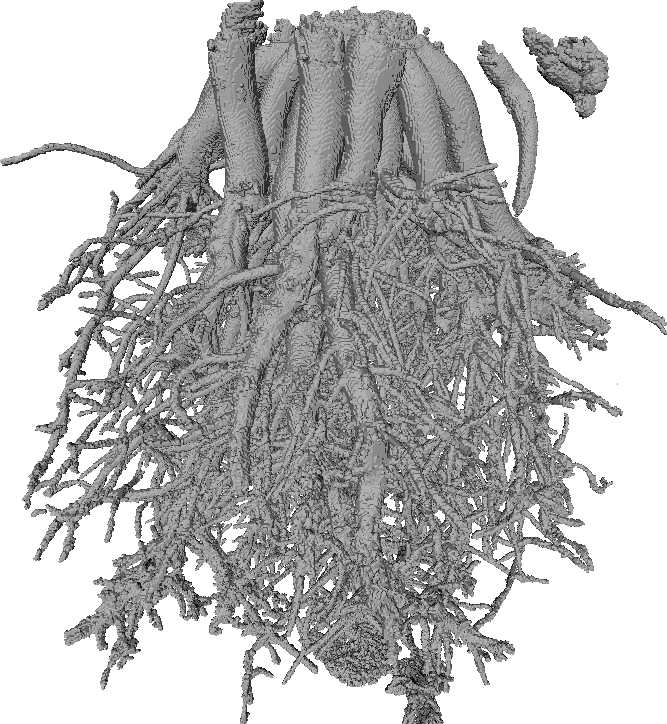}
}
\subfloat[Overhead view] 
{\label{sf:maize1}
\includegraphics[width=0.5\linewidth]{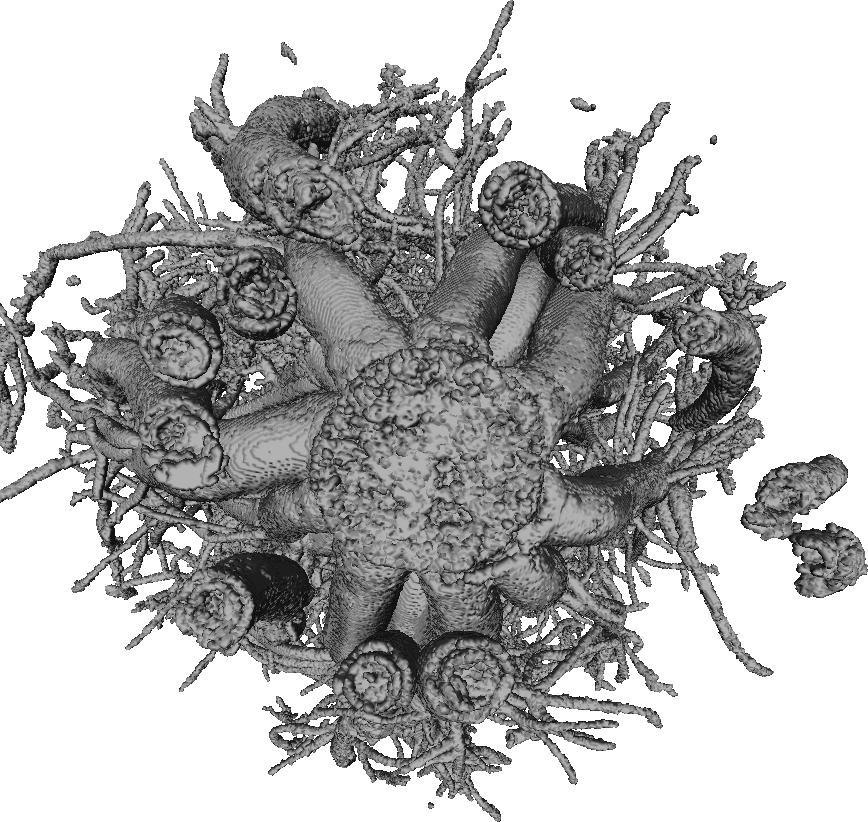}
}

\subfloat[Detail of root structure] 
{ \label{sf:maize2}
\includegraphics[width=0.8\linewidth]{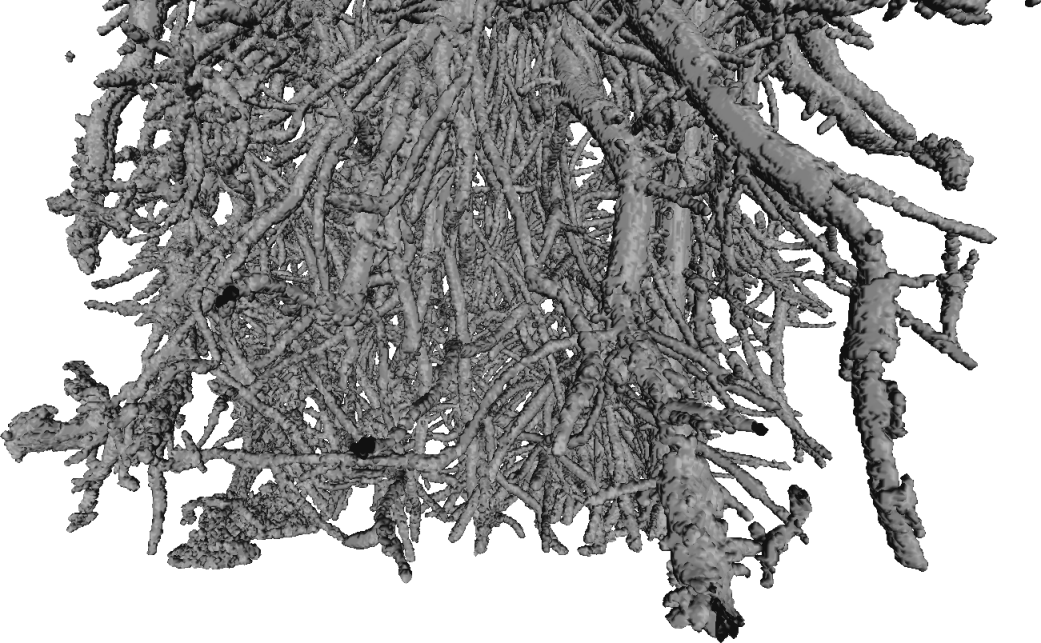}
}
\caption{Results of using the proposed method with Dataset 1.  The species is maize, and only a portion of the pot was acquired of this dataset.  It is because of this that roots are terminated along a plane, for instance in \ref{sf:maize2}. The structures to the right in \ref{sf:maize1} are not false positives, but rather crown roots.  All of the three-dimensional models in the this paper consist of isosurfaces generated by BoneJ \cite{bonej} and were visualized using Meshlab \cite{meshlab}.  }
\label{fig:dataset1}
\end{figure}

\begin{figure}[]
\centering
\includegraphics[width=0.84\linewidth]{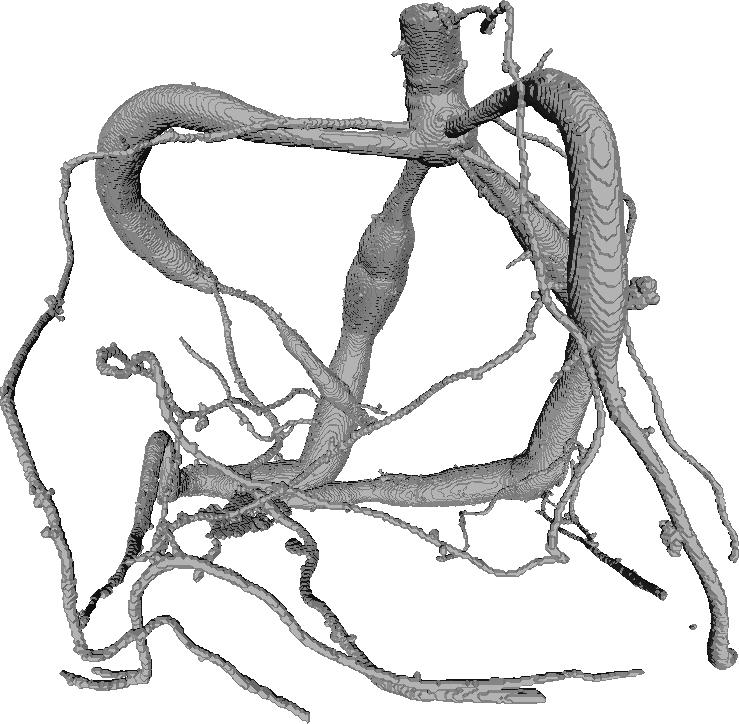}
\caption{Results of using the proposed method with Dataset 2.  The species is cassava, with a mixture of large and small, thin roots.}
\label{fig:dataset2}
\end{figure}

\begin{figure}[]
\centering
\subfloat[] 
{
\includegraphics[angle =90, width=1\linewidth]{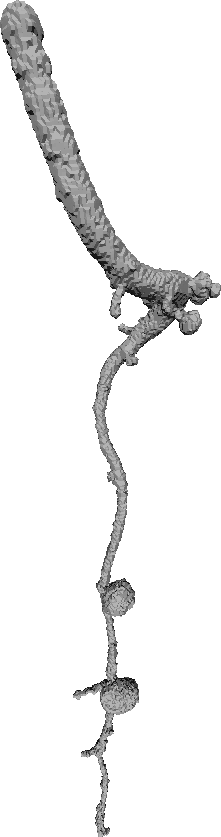}
}

\subfloat[] 
{
\includegraphics[angle = 90, width=1\linewidth]{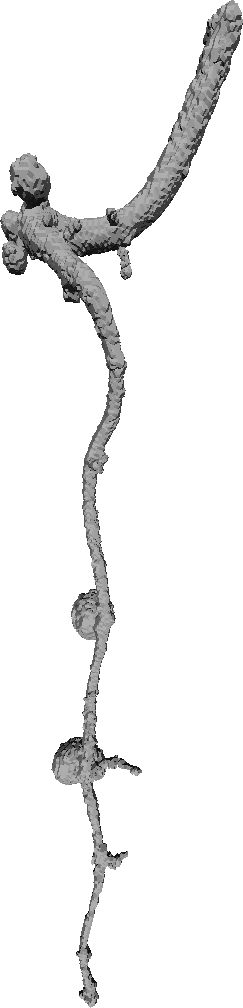}
}
\caption{Results of using the proposed method with Dataset 3.  The species is Soybean.  The rounded regions are nitrogen-fixing nodules characteristic of soybean and other legumes; this root system is rotated ninety degrees for space considerations.}
\label{fig:dataset3}
\end{figure}

\begin{figure}[]
\centering
\subfloat[Side view] 
{
\includegraphics[width=0.62\linewidth]{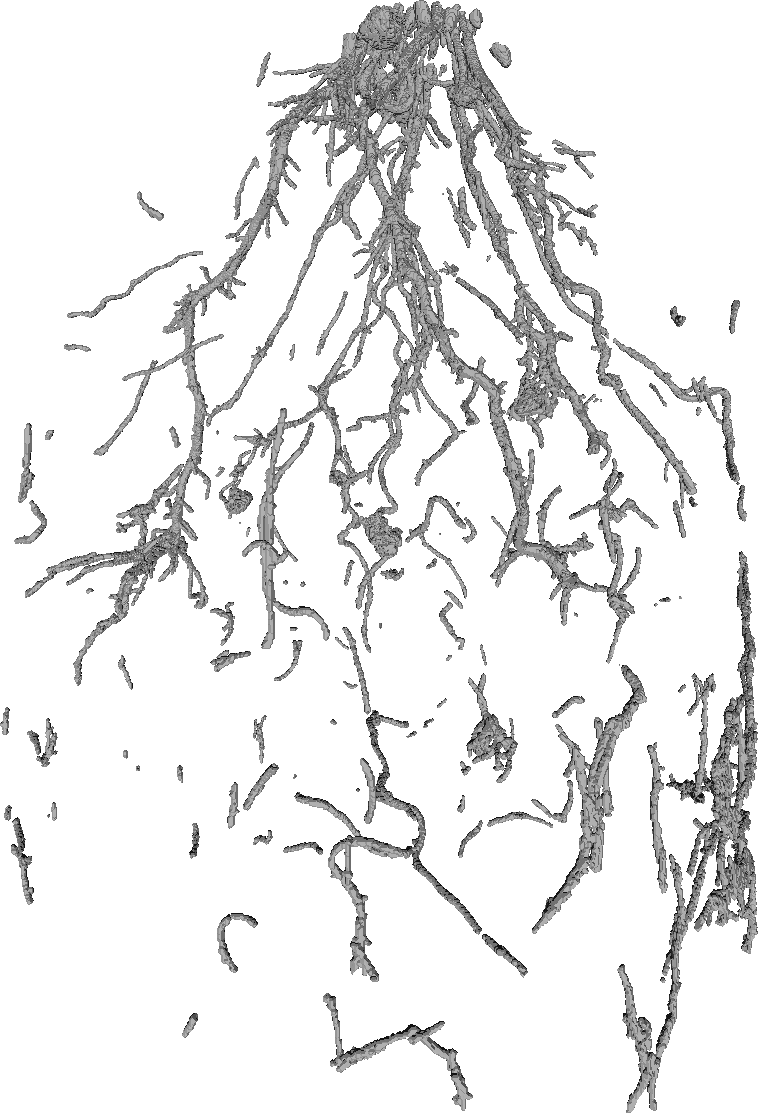}
}

\subfloat[Detail]
{  \label{sf:maizedetail}
\includegraphics[width=0.6\linewidth]{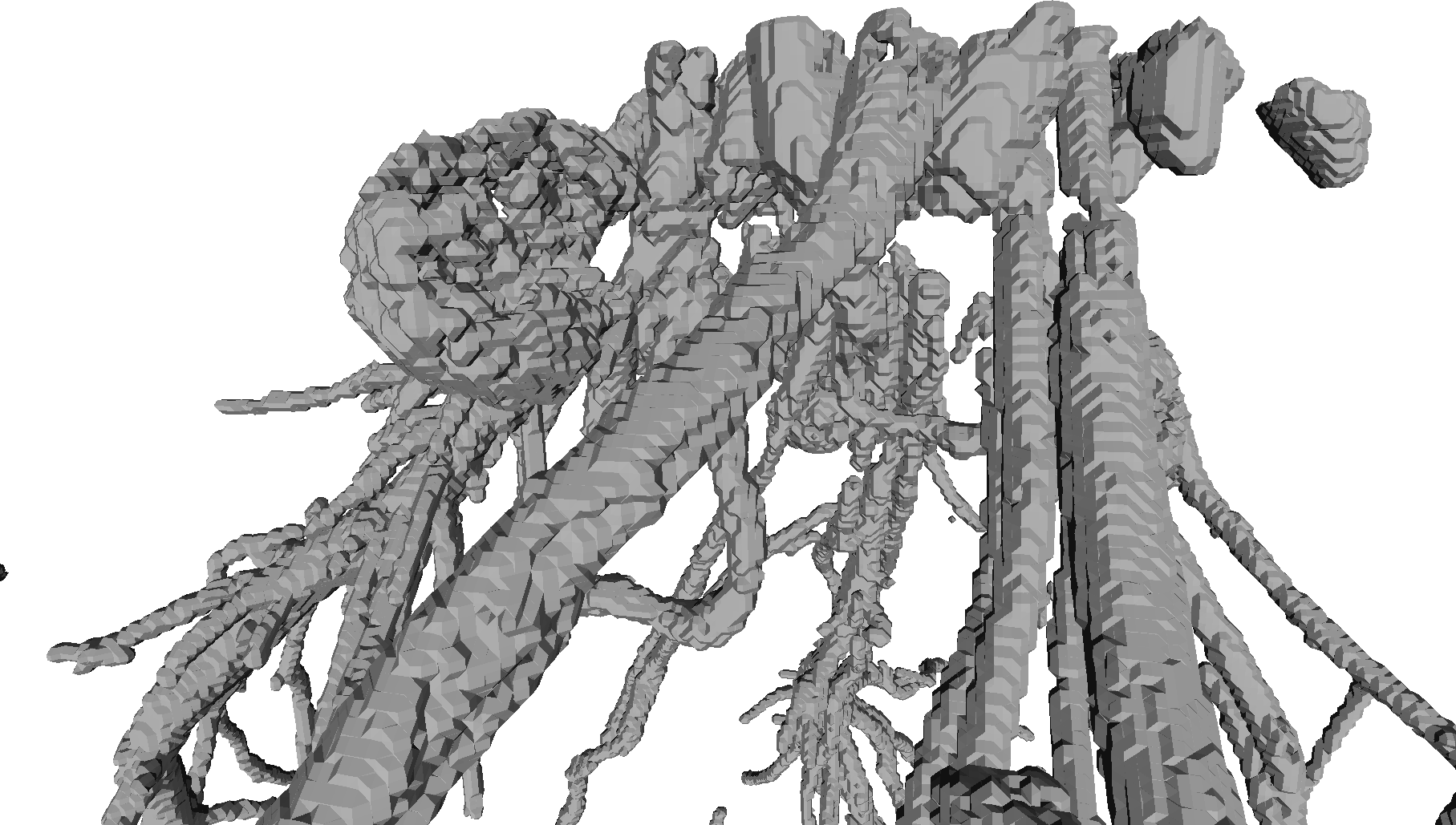}
} 
\caption{Results of using the proposed method with Dataset 4.  The species is maize. \ref{sf:maizedetail} shows the `leaking' from root regions to non-root regions, in this case, an expanded clay particle, that occurs during the execution of the method on this dataset.  }
\label{fig:dataset4}
\end{figure}

\begin{figure}[]
\centering
\includegraphics[width=0.6\linewidth]{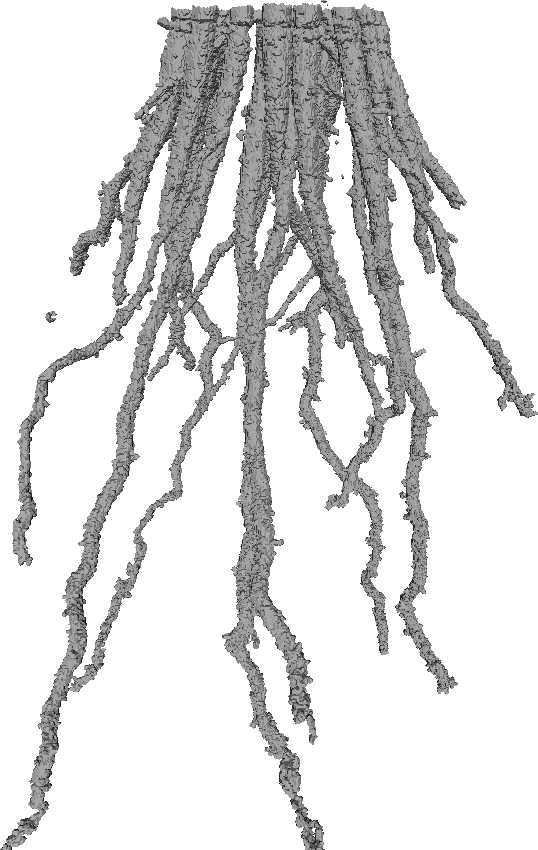}
\caption{Results of using the proposed method with Dataset 5.  The species is maize.}
\label{fig:dataset5}
\end{figure}

The datasets represent a range of characteristics in terms of the texture and greylevels of the medium as well as root thickness, as defined in terms of pixels.  The reconstruction results of using the proposed method are shown in Figures \ref{fig:dataset1}-\ref{fig:dataset5}.  The maize plant from Dataset 1 and shown in Figure \ref{fig:dataset1} represents the most mature root system of the set.  Qualitatively, the method performs well here, recovering both the large and small features of the root system.  False positives are present mainly on the lower side of the plant, where some artifacts in imaging resulted in blurring near the edges.

The cassava root system from Dataset 2 (Figure \ref{fig:dataset2}) is also characterized by large and small roots, but with a different hierarchy than maize.  We found that in this dataset, better results were obtained when the iterative growing adaption in Section \ref{ss:g-h} was not used; as a result, we set $G_h\left(y\right) = 1$ for all $y$ at the start of the method.  It may be that larger regions than $b$ are needed to model $\Omega_1$, since the root regions are so large in Dataset 2 compared to other datasets.  The method was able to recover very small roots which were not obvious to the human eye when viewing the dataset slice by slice, and reconstructs the large roots well. 

Dataset 3 is the only example of soybean in our experiments, and that species' root system is characterized by the traditional thin, elongated structures as well as rounded regions called nodules.  The $\nu$ parameter was increased to $1.2$ in this dataset to prevent leaking from the nodules, in the $\Omega_2$ class, to the interior of the expanded clay particles, which had similar greylevels.  Very small roots were not recovered, and in the images the appearance of the very small roots was very faint.  However, larger roots and all of the nodules were recovered. Unusually, this dataset required only one initialization image and had a reasonable run time, at eight minutes.    

Dataset 4 was an example of maize planted in expanded clay.  Here, the small size of the roots (defined by the number of pixels) in the middle to lower sections of the pot combined with the expanded clay medium resulted in some challenges for the proposed method.  Figure \ref{fig:dataset4} shows the recovered root system. As is evident, sections are recovered but the root system is much more extensive than was discovered by the method.  In addition, in this dataset alteration of the $\nu$ parameter did not prevent the leaking phenomenon into expanded clay particles, shown in Figure \ref{sf:maizedetail} near the top of the root system.  

The last dataset again shows an example of maize, and this time the medium was turface, and is shown in Figure \ref{fig:dataset5}.  In this dataset, the differentiation of medium from root can be quite difficult visually, and the root regions are very noisy.  The method was able to recover parts of the root system without false positives, however the root reconstruction was truncated and smaller roots not reconstructed.

\begin{figure}[]
\centering
\subfloat[Original image] 
{
\includegraphics[width=0.48\linewidth]{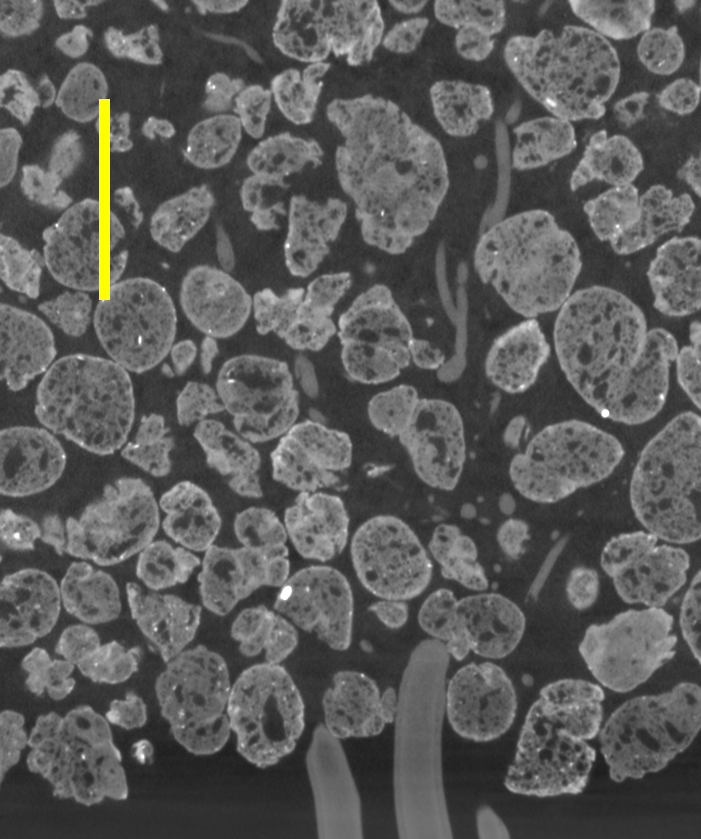}
}
\subfloat[Results overlaid on image] 
{
\includegraphics[width=0.48\linewidth]{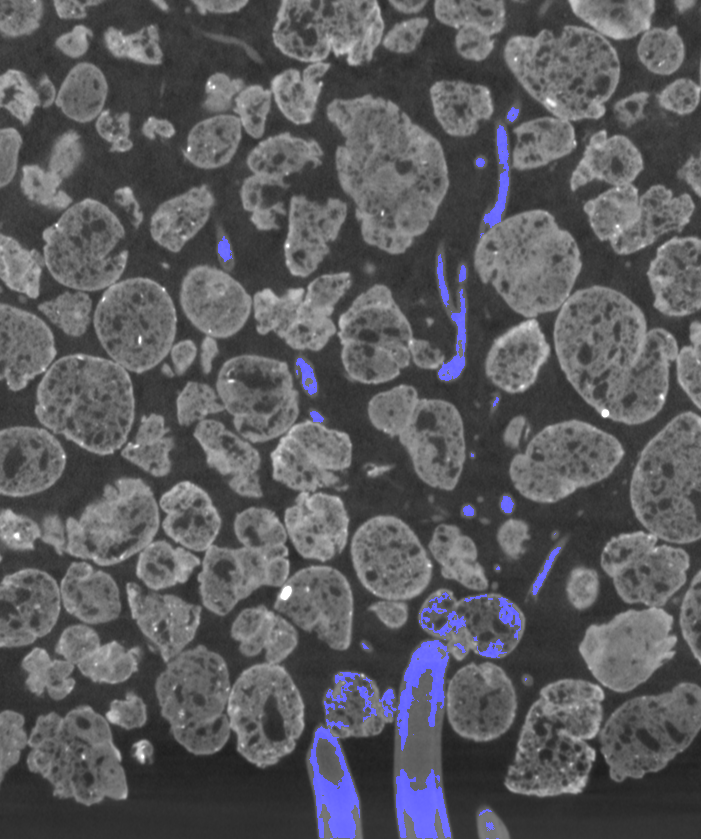}
}
\caption{Detail of an image from dataset 4, showing the variability in the size of the roots as well as greylevels.  On the left, the yellow bar indicates a region of $200$ pixels' length.  On the right, areas in blue color indicate areas marked as root regions by the method.}
\label{fig:dataset4hardexample}
\end{figure}

The method was able to recover larger roots on the majority of datasets, with few false positives.  Small root recovery was problematic, though that may be a function of the parameters of the X-ray CT data acquisition.  Figure \ref{fig:dataset4hardexample} shows, as an example, the scale of the small roots in Dataset 4 and the corresponding detection by the proposed method.  
\begin{figure}[]
\centering
\includegraphics[width = 1.0\linewidth]{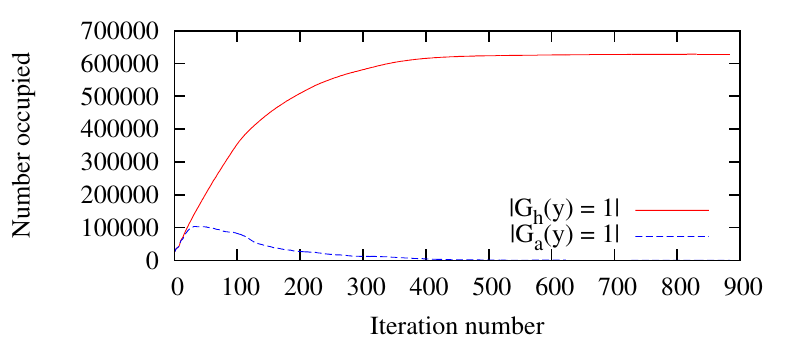}
\caption{Graph of the number of occupied voxels in the occupancy grids $G_h$ and $G_a$ for Dataset 1, by iteration.  }
\label{fig:gs_convergence}
\end{figure}

The use of active regions of the contour resulted in a particular pattern concerning the number of active regions ($G_a(y) = 1$) versus the number of regions explored so far ($G_h(y) = 1$).  Figure \ref{fig:gs_convergence} shows a graph of these two values over all iterations for Dataset 1.  Consequently, the computational resources are focused on smaller and smaller regions near the end of the algorithm, as the total number of regions to be explored nears convergence.  An illustration of the active regions on a reconstruction is shown in Figure \ref{fig:dataset2composite}.

\subsubsection{Comparison to other approaches}

\begin{figure}[]
\centering
\includegraphics[angle = 90, width = 0.2\linewidth]{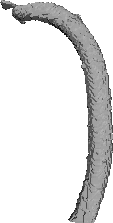}
\caption{Segmentation of Dataset 3 using the RooTrak method \cite{Mairhofer2013Recovering, Mairhofer2012Rootrak}.  The method terminates at image slice $345$, when the first nodules are encountered. This result rotated ninety degrees for space considerations. }
\label{fig:rootrak}
\end{figure}

We did asses the performance of another approach, RooTrak \cite{Mairhofer2013Recovering, Mairhofer2012Rootrak}, on two of the datasets here, Datasets 2 and 3.  In RooTrak, there is a similarity measure enforced between neighboring layers.  We found in both of these cases that RooTrak terminated early and hypothesize that the reason is that the size of the root regions between layers differed by too great a value in the types of situations that we consider; for Dataset 2, the maximum number of iterations was $11$.  For Dataset 3, the result was better with the maximum number of iterations at $345$, but termination was still very early.  The result is shown in Figure \ref{fig:rootrak}.  The implementations of another method, the Root1 plugin for ImageJ \cite{Flavel2017image} is not available at this time.    

\section{Conclusions}

We presented a methodology for segmenting plant roots from X-ray CT images of them growing in soil.  The method is intended to be used for a variety of media and plant root maturity levels and sizes.  

While this was a general method, some incorporation of features of particular media might be useful to extracting better segmentations.  For instance, in the expanded clay medium sometimes the front propagation produces a leak from the root to a section of an expanded clay granule because of the similar greylevels between the two; we mitigate this situation by increasing the value of the smoothness parameter $\nu$.  However, if the imaging of plant roots in a particular medium is necessary on a large scale for an experiment, modifying this method by considering specialized solutions to removing known non-root shapes given the plant species, X-ray CT imaging parameters, root size, and medium may be more efficient.  

In addition, while the region exploration adaption in Section \ref{ss:mod} reduces some issues associated with regional variation of texture and greylevels, it may be that distributions may need to be defined regionally to take into account physical realities.


{\small
\bibliographystyle{ieee}
\bibliography{level_sets}

\begin{thebibliography}{10}\itemsep=-1pt

\bibitem{Abiven2015}
S.~Abiven, A.~Hund, V.~Martinsen, and G.~Cornelissen.
\newblock Biochar amendment increases maize root surface areas and branching: a
  shovelomics study in zambia.
\newblock {\em Plant and Soil}, 395(1):45--55, Oct 2015.

\bibitem{Chowdhury2017cortical}
M.~Chowdhury, D.~J{\"o}rgens, C.~Wang, {\"O}.~Smedby, and R.~Moreno.
\newblock Segmentation of cortical bone using fast level sets.
\newblock In {\em Proc. SPIE}, volume 10133, pages 1013327--1 -- 1013327--7,
  2017.

\bibitem{meshlab}
P.~Cignoni, M.~Callieri, M.~Corsini, M.~Dellepiane, F.~Ganovelli, and
  G.~Ranzuglia.
\newblock {MeshLab: an Open-Source Mesh Processing Tool}.
\newblock In V.~Scarano, R.~D. Chiara, and U.~Erra, editors, {\em Eurographics
  Italian Chapter Conference}. The Eurographics Association, 2008.

\bibitem{clark2011three}
R.~T. Clark, R.~B. MacCurdy, J.~K. Jung, J.~E. Shaff, S.~R. McCouch, D.~J.
  Aneshansley, and L.~V. Kochian.
\newblock Three-dimensional root phenotyping with a novel imaging and software
  platform.
\newblock {\em Plant Physiology}, 156(2):455--465, 2011.

\bibitem{Colombi2015}
T.~Colombi, N.~Kirchgessner, C.~A. Le~Mari{\'e}, L.~M. York, J.~P. Lynch, and
  A.~Hund.
\newblock Next generation shovelomics: set up a tent and rest.
\newblock {\em Plant and Soil}, 388(1):1--20, Mar 2015.

\bibitem{Cremers2007}
D.~Cremers, M.~Rousson, and R.~Deriche.
\newblock A review of statistical approaches to level set segmentation:
  Integrating color, texture, motion and shape.
\newblock {\em International Journal of Computer Vision}, 72(2):195--215, Apr
  2007.

\bibitem{bonej}
M.~Doube, M.~M. Kłosowski, I.~Arganda-Carreras, F.~P. Cordelières, R.~P.
  Dougherty, J.~S. Jackson, B.~Schmid, J.~R. Hutchinson, and S.~J. Shefelbine.
\newblock Bonej: Free and extensible bone image analysis in imagej.
\newblock {\em Bone}, 47(6):1076 -- 1079, 2010.

\bibitem{Farag2013novel}
A.~A. Farag, H.~E. A.~E. Munim, J.~H. Graham, and A.~A. Farag.
\newblock A novel approach for lung nodules segmentation in chest ct using
  level sets.
\newblock {\em IEEE Transactions on Image Processing}, 22(12):5202--5213, Dec
  2013.

\bibitem{Flavel2017image}
R.~J. Flavel, C.~N. Guppy, S.~M.~R. Rabbi, and I.~M. Young.
\newblock An image processing and analysis tool for identifying and analysing
  complex plant root systems in 3d soil using non-destructive analysis: Root1.
\newblock {\em PLOS ONE}, 12(5):1--18, 05 2017.

\bibitem{Ghadimi2016skull}
S.~Ghadimi, H.~A. Moghaddam, R.~Grebe, and F.~Wallois.
\newblock Skull segmentation and reconstruction from newborn ct images using
  coupled level sets.
\newblock {\em IEEE Journal of Biomedical and Health Informatics},
  20(2):563--573, March 2016.

\bibitem{lefohn2004streaming}
A.~E. Lefohn, J.~M. Kniss, C.~D. Hansen, and R.~T. Whitaker.
\newblock A streaming narrow-band algorithm: interactive computation and
  visualization of level sets.
\newblock {\em IEEE Transactions on Visualization and Computer Graphics},
  10(4):422--433, July 2004.

\bibitem{Mairhofer2016Visual}
S.~Mairhofer, J.~Johnson, C.~J. Sturrock, M.~J. Bennett, S.~J. Mooney, and
  T.~P. Pridmore.
\newblock Visual tracking for the recovery of multiple interacting plant root
  systems from x-ray $\mu$ ct images.
\newblock {\em Machine Vision and Applications}, 27(5):721--734, Jul 2016.

\bibitem{Mairhofer2013Recovering}
S.~Mairhofer, S.~Zappala, S.~Tracy, C.~Sturrock, M.~J. Bennett, S.~J. Mooney,
  and T.~P. Pridmore.
\newblock Recovering complete plant root system architectures from soil via
  x-ray $\mu$-computed tomography.
\newblock {\em Plant Methods}, 9(1):8, Mar 2013.

\bibitem{Mairhofer2012Rootrak}
S.~Mairhofer, S.~Zappala, S.~R. Tracy, C.~Sturrock, M.~Bennett, S.~J. Mooney,
  and T.~Pridmore.
\newblock Rootrak: Automated recovery of three-dimensional plant root
  architecture in soil from x-ray microcomputed tomography images using visual
  tracking.
\newblock {\em Plant Physiology}, 158(2):561--569, 2012.

\bibitem{meijster2002general}
A.~Meijster, J.~B. Roerdink, and W.~H. Hesselink.
\newblock A general algorithm for computing distance transforms in linear time.
\newblock In {\em Mathematical Morphology and its applications to image and
  signal processing}, pages 331--340. Springer, 2002.

\bibitem{Metzner2015direct}
R.~Metzner, A.~Eggert, D.~van Dusschoten, D.~Pflugfelder, S.~Gerth, U.~Schurr,
  N.~Uhlmann, and S.~Jahnke.
\newblock Direct comparison of mri and x-ray ct technologies for 3d imaging of
  root systems in soil: potential and challenges for root trait quantification.
\newblock {\em Plant Methods}, 11(1):17, Mar 2015.

\bibitem{pineros2016evolving}
M.~A. Pi{\~{n}}eros, B.~G. Larson, J.~E. Shaff, D.~J. Schneider, A.~X. Falcão,
  L.~Yuan, R.~T. Clark, E.~J. Craft, T.~W. Davis, P.-L. Pradier, N.~M. Shaw,
  I.~Assaranurak, S.~R. McCouch, C.~Sturrock, M.~Bennett, and L.~V. Kochian.
\newblock Evolving technologies for growing, imaging and analyzing 3d root
  system architecture of crop plants.
\newblock {\em Journal of Integrative Plant Biology}, 58(3):230--241, 2016.

\bibitem{Rousson2002Variational_TR}
M.~Rousson and R.~Deriche.
\newblock A variational framework for active and adaptative segmentation of
  vector valued images.
\newblock Technical Report 4515, INRIA Sophia Antipolis, July 2002.

\bibitem{Rousson2002Variational_WS}
M.~Rousson and R.~Deriche.
\newblock A variational framework for active and adaptative segmentation of
  vector valued images.
\newblock In {\em Workshop on Motion and Video Computing, 2002. Proceedings.},
  pages 56--61, Dec 2002.

\bibitem{Shahzad2017automated}
R.~Shahzad, S.~Gao, Q.~Tao, O.~Dzyubachyk, and R.~van der Geest.
\newblock {\em Automated Cardiovascular Segmentation in Patients with
  Congenital Heart Disease from 3D CMR Scans: Combining Multi-atlases and
  Level-Sets}, pages 147--155.
\newblock Springer International Publishing, Cham, 2017.

\bibitem{Symonova2015dynamic}
O.~Symonova, C.~N. Topp, and H.~Edelsbrunner.
\newblock Dynamicroots: A software platform for the reconstruction and analysis
  of growing plant roots.
\newblock {\em PLOS ONE}, 10(6):1--15, 06 2015.

\bibitem{Topp2013-3D}
C.~N. Topp, A.~S. Iyer-Pascuzzi, J.~T. Anderson, C.-R. Lee, P.~R. Zurek,
  O.~Symonova, Y.~Zheng, A.~Bucksch, Y.~Mileyko, T.~Galkovskyi, B.~T. Moore,
  J.~Harer, H.~Edelsbrunner, T.~Mitchell-Olds, J.~S. Weitz, and P.~N. Benfey.
\newblock 3d phenotyping and quantitative trait locus mapping identify core
  regions of the rice genome controlling root architecture.
\newblock {\em Proceedings of the National Academy of Sciences},
  110(18):E1695--E1704, 2013.

\bibitem{Trachsel2011}
S.~Trachsel, S.~M. Kaeppler, K.~M. Brown, and J.~P. Lynch.
\newblock Shovelomics: high throughput phenotyping of maize (zea mays l.) root
  architecture in the field.
\newblock {\em Plant and Soil}, 341(1):75--87, Apr 2011.

\bibitem{Wang2011level}
C.~Wang, H.~Frimmel, and {\"O}.~Smedby.
\newblock Level-set based vessel segmentation accelerated with periodic
  monotonic speed function.
\newblock In {\em In Proc. SPIE}, volume 7962, pages 7962 -- 7962 -- 7, 2011.

\bibitem{Wang2014fast}
C.~Wang, H.~Frimmel, and {\"O}.~Smedby.
\newblock Fast level-set based image segmentation using coherent propagation.
\newblock {\em Medical Physics}, 41(7):073501--n/a, 2014.
\newblock 073501.

\bibitem{Whitaker2001variable}
R.~T. Whitaker and X.~Xue.
\newblock Variable-conductance, level-set curvature for image denoising.
\newblock In {\em Proceedings 2001 International Conference on Image Processing
  (Cat. No.01CH37205)}, volume~3, pages 142--145 vol.3, 2001.

\bibitem{Yang2017left}
C.~Yang, W.~Wu, Y.~Su, and S.~Zhang.
\newblock Left ventricle segmentation via two-layer level sets with circular
  shape constraint.
\newblock {\em Magnetic Resonance Imaging}, 38(Supplement C):202 -- 213, 2017.

\bibitem{Zheng2011Detailed}
Y.~Zheng, S.~Gu, H.~Edelsbrunner, C.~Tomasi, and P.~Benfey.
\newblock Detailed reconstruction of 3d plant root shape.
\newblock In {\em 2011 International Conference on Computer Vision}, pages
  2026--2033, Nov 2011.

\end{thebibliography}
}

\end{document}